\documentclass[sn-mathphys]{sn-jnl}

\jyear{2021}%

\theoremstyle{thmstyleone}%
\PassOptionsToPackage{hyphens}{url}\usepackage{hyperref}
\theoremstyle{thmstyletwo}%

\theoremstyle{thmstylethree}%
\newtheorem{definition}{Definition}%
\usepackage{hyperref}
\usepackage{xcolor}

\newcommand{\Dgm}{\mathrm{Dgm}}
\newcommand{\im}{\mathrm{im}}
\newcommand{\Ent}{\mathrm{Ent}}

\usepackage{natbib}
\usepackage{algorithm}

\begin{document}

\title[Summary and Distance between  Sets of Texts based on TDA]{Summary and Distance between Sets of Texts based on Topological Data Analysis
}


\author[1]{\fnm{Rocio} \sur{Gonzalez-Diaz}}\email{rogodi@us.es}
\equalcont{These authors contributed equally to this work.}

\author[2]{\fnm{Miguel A.} \sur{Gutiérrez-Naranjo}}\email{magutier@us.es}
\equalcont{These authors contributed equally to this work.}

\author*[1]{\fnm{Eduardo} \sur{Paluzo-Hidalgo}}\email{paluzo@us.es}
\equalcont{These authors contributed equally to this work.}

\affil[1]{\orgdiv{Dept. of Applied Mathematics I}, \orgname{University of Sevilla}, \orgaddress{\city{Sevilla}, \postcode{41012}, \country{Spain}}}

\affil[2]{\orgdiv{Dept. of Computer Sciences and Artificial Intelligence}, \orgname{University of Sevilla}, \orgaddress{\city{Sevilla}, \postcode{41012}, \country{Spain}}}


\abstract{In this paper, we use topological data analysis (TDA) tools as persistent homology, persistent entropy and bottleneck distance, to provide a {\it TDA-based summary} of any given set of texts and a general method for computing a distance between any two literary styles, authors or periods. 
To this aim, deep-learning word-embedding techniques are combined with these tools in order to study topological properties of texts embedded in a metric space.
As a case of study, we use the written texts of three poets of the Spanish Golden Age: Francisco de Quevedo, Luis de G\'ongora and Lope de Vega. 
As far as we know, this is the first time that word embedding, bottleneck distance, persistent homology and persistent entropy are used together to characterize texts and to compare different literary styles.}




\maketitle
\section{Introduction}
Topology is the branch of mathematics which deals with proximity relations and continuous deformations of abstract spaces. 
Recently, many researchers have paid attention to it due to the increasing amount of data available and the need for in-depth analysis of these datasets to extract useful properties of the space they sample. 
The application of topological tools to the study of data, known as Topological Data Analysis (TDA), has achieved a long list of successes in recent years (see, e.g., \citep{DBLP:journals/corr/abs-1907-08325},  \citep{DBLP:journals/corr/abs-1904-02971} or \citep{DBLP:conf/icml/RamamurthyVM19}, among many others). 
In this paper, we focus our attention on applying TDA techniques to study and effectively compute a kind of {\it proximity} among literary styles.

Until now, most of the methods used in comparative studies in philology are essentially qualitative. 
The comparison among writers, periods or, in general, literary styles is often based on stylistic analysis that cannot be quantified.  
Several quantitative methods in linguistics were applied in the past (see \citep{johnson2008quantitative}) but their use is still controversial \citep{Rahman2017}.

Our aim is to provide quantitative methods to classify texts, authors and literary styles, in general. 
But, instead of only using  statistical methods, our procedure is based on the analysis of the spatial shape of the data after embedding it in a high-dimensional metric space. Broadly speaking, we start by representing a text as a cloud of points by using the so-called {\it word embedding} technique.
The second key point of our method is the use of some TDA techniques, such as the persistent entropy and the bottleneck distance, to measure the proximity between different  point clouds  representing different texts, authors, literary styles, etc. 
The reason why we use persistent entropy is that it is a summary tool easy to compute and stable under small changes in the input data  \cite{ATIENZA2020107509}.

Let us recall that word embedding techniques try to find a representation of a set of words on a given alphabet
as a high-dimensional point cloud in such a way that the semantic proximity is kept.
Among the most popular systems for word embedding, 
the \texttt{word2vec} \citep{DBLP:journals/corr/MikolovLS13}, GloVe \citep{DBLP:conf/emnlp/PenningtonSM14} or FastText \citep{DBLP:journals/tacl/BojanowskiGJM17} systems can be cited.
Along this paper, the \texttt{word2vec} system with its \texttt{skipgram} variation will be used to obtain said multidimensional representation of the texts.

Regarding the  study of proximity  between the word embedding of different texts, there are many  ways in computer science to study dissimilarities and to measure the  distance between two point clouds \citep{deza2009encyclopedia}, but most of them are merely based on some kind of statistical summary of the point cloud and not on its {\it shape}. 
TDA techniques can capture the structure representation of data distribution, as shown in \cite{6373736} and hence, it can be considered a powerful tool in combination with 
machine learning to be used in the different areas of data analysis (see, for example, \cite{DBLP:journals/air/ZielinskiLJZD21} and \cite{dey2020topo}).
In spite of the doubtless interest in quantifying and measuring the proximity between
literary styles, as far as we know, very few researchers explored the dissimilarity and proximity between them by combining machine learning and TDA techniques (see, for example, \citep{bdcc2040033,teminas2018local, Wright_Zheng_2020}). 

Our contribution is two-folded. 
Firstly, to the best of our knowledge, this is the first time that persistent entropy is applied to language processing problems. 
Secondly, the concept of a {\it TDA-based summary} of a set of texts is introduced.
Specifically, the shape of a point cloud representing  a text is captured by using a TDA technique known as persistence diagrams, which is based on deep and well-known concepts of algebraic topology such as simplicial complexes, homology groups and filtrations. Specifically, to summarize a persistence diagram, we will compute its persistent entropy. 
Persistent entropy is based on the Shannon entropy and it has been successfully applied in many real-world problems such as to characterize  epithelial tissues images \cite{10.1007/978-3-030-10828-1_14} or to measure the heart-rate   variability to a  sleep-wake classification \cite{chung2020persistence}.
Persistent entropy  will be applied to provide a TDA-based summary of a set of texts and to characterize the literary works made by an author. 
A distance  between persistence diagrams, namely the bottleneck distance, provides a way to quantify the proximity between  two different persistence diagrams and, hence, a way to quantify the proximity between  two different literary styles.

In order to illustrate the potential of  the proposed technique, we provide a comparison of the literary works of two poets, Luis de G\'ongora and Francisco de Quevedo, who are representatives of the two main Spanish Golden Age literary styles
called {\it Culteranismo} and  {\it Conceptismo}, respectively.
We also consider a third poet, called Lope de Vega.
Literary experts agree that Lope de Vega and Francisco de Quevedo styles are {\it close} (they both belong to Conceptismo), but both are {\it far} from the style of Luis de G\'ongora, which belongs to Culteranismo \citep{rutherford2016spanish}. 
The application of TDA techniques made in this paper for measuring the proximity between such literary styles, quantitatively confirms that the styles of Lope de Vega and Francisco de Quevedo are close to each other and yet both are far from the style of Luis~de~G\'ongora.

The paper is organized as follows: In Section \ref{sec:background}, some preliminary notions about word embedding and TDA techniques are provided. The procedure applied to compute a TDA-based summary of a set of texts and to compare two different literature styles is described in Section \ref{sec:methodology}. In Section \ref{sec:experiment}, the specific computation of TDA-based summary of the literary works of each of the three poets mentioned above, and a comparison between their literary styles
is thoroughly described. 
Finally, in Section \ref{sec:conclusions}, conclusions and future work are given.

\section{Background}\label{sec:background}
In this section we recall some basics related to the techniques used along the paper.
Firstly, word embedding methodology is briefly introduced. Later, the relevant tools from TDA used in our approach are described.

\subsection{Word Embedding}\label{sec:background_word2vec}

Word embedding is the collective name of a set of methods for representing words from natural languages as points (or vectors) in a real-valued multi-dimensional space. 
The common feature of such methods is that words with similar meanings take close representation. 
Such representation methods are on the basis of some of the big successes of deep learning applied to natural language processing (see, for example, \citep{DBLP:conf/nips/YinS18} or \citep{DBLP:journals/corr/abs-1901-09069}). 
Next, we recall some basic definitions related to this methodology.

\begin{definition}[corpus]
Given a set of words on a given alphabet, a {\it corpus} is a finite collection of writings composed with these words,
denoted by $C$. 
The vocabulary, $V$, of a corpus $C$ is the set of all the words that appear in $C$. 
Finally, given $d\ge 1$, a word embedding is a function $E:V\rightarrow \mathbb{R}^d$.
\end{definition}

\begin{figure}
    \centering
    \includegraphics[width =0.6 \linewidth]{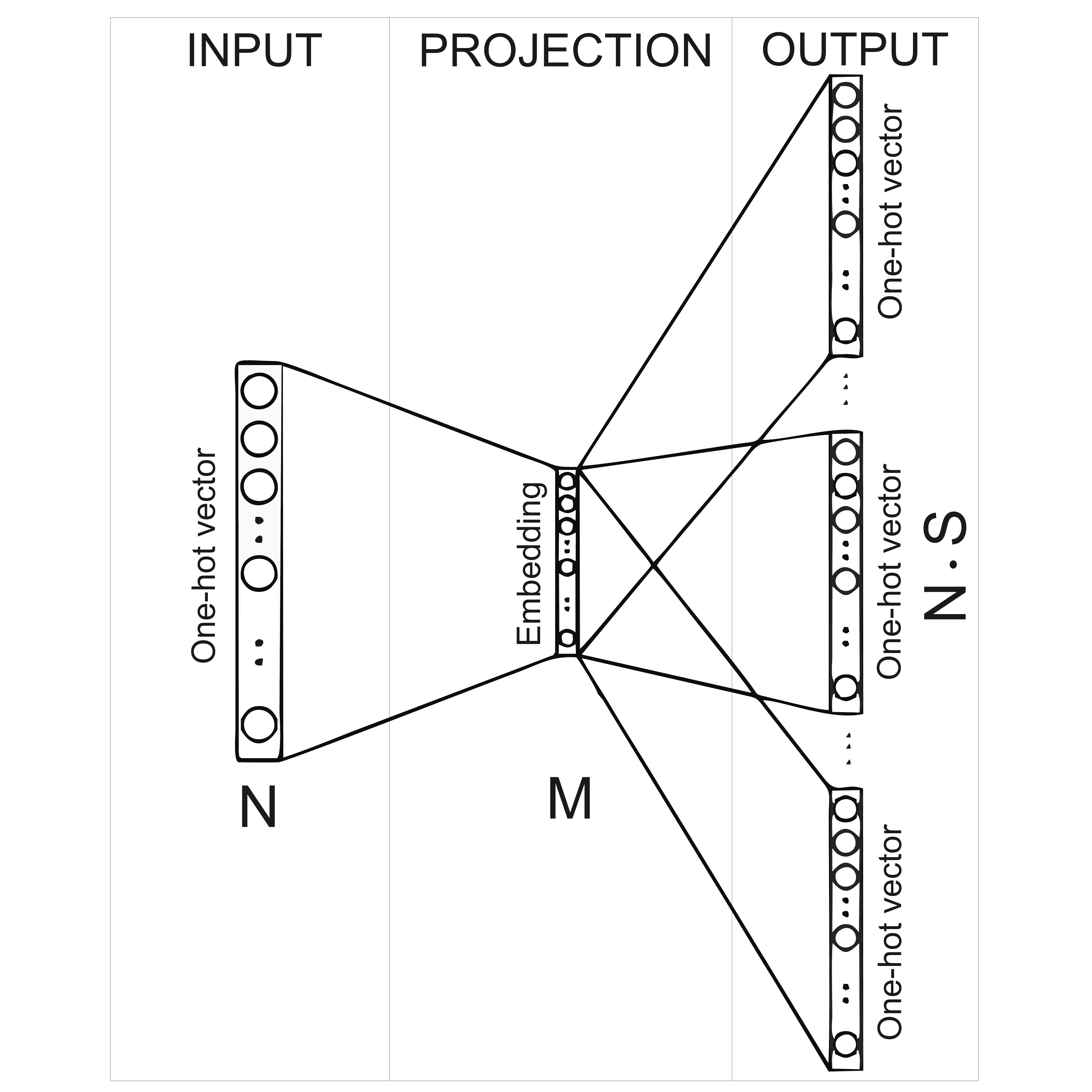}
    \caption{The \texttt{skipgram} neural network architecture. The input layer has as many neurons as the length of the one-hot vector that encode the words of the corpus, i.e.,
    the number of words 
    that compose the vocabulary of the corpus, $N$ in this case. The size of the projection layer 
    is equal to the dimension in which we want to embed the corpus, $M$. Finally, the output layer has $N\cdot S$ neurons where $S$ is the size of the window, i.e., the number of surrounding words that the model tries to predict. This image is inspired in the image of the \texttt{skipgram} model in \citep{skipgram}.}
    \label{fig:skipgram}
\end{figure}

The word embedding process used along this paper is the \texttt{word2vec}\footnote{The model we used is the one implemented in the python library \texttt{gensim} which is based on \citep{Mikolov2013EfficientEO,Mikolov:2013:DRW:2999792.2999959}.}, specifically, its modified version called \texttt{skipgram} \citep{DBLP:conf/lrec/GuthrieA0GW06}. 
It is based on a neural network architecture with one hidden layer where the input is a corpus and the output is a probability distribution. 
We train it with a corpus to detect similarities in words based on their relative distance in a writing. 
Such distance is the base of their representation in an $n$-dimensional space.

Roughly speaking, the \texttt{skipgram} neural network is trained by using a corpus, where the context of a word is considered as a {\it window} around a target word. 
In this way, in the \texttt{skipgram} model, each word of the input is processed by a log-linear classifier with continuous projection layer, trying to predict the previous and the following words in a sentence. 
In this kind of neural network architecture, the input is a one-hot vector representing a word of the corpus 
and the output is a prediction of the surrounding words. 
More specifically, the neural network
follows the architecture shown and explained in Figure~\ref{fig:skipgram}.


\subsection{Topological Data Analysis}

The field of computational topology and topological data analysis were born as a  combination of topics in geometry, topology, and algorithmics. 
In this section, some of their basic concepts are recalled. For a a detailed presentation of these fields, \citep{Edelsbrunner10} and \cite{carlsson_vejdemo-johansson_2021} are recommended. 

We will recall, firstly, homology and, lately, persistent homology as fundamental tools of TDA.
The information obtained when computing persistent homology is usually encapsulated in a persistence diagram. 
Next, the concept of persistent entropy is introduced as a summary tool for persistence diagrams. 
Finally, the bottleneck distance will be shown as the main distance to compare persistence diagrams.

The class of the spaces where we define homology groups are the underlying spaces of simplicial complexes which are combinatorial structures built from lines, segments, triangles, and so on for higher dimensions. 
These components are called simplices.

\begin{definition}[$n$-simplex]
Let $\{v_0,\dots , v_n\}$ be a set of geometrically independent points in $\mathbb{R}^d$. 
The $n$-simplex $\sigma$ spanned by $v_0,\dots, v_n$ 
is defined as the set of all points $x\in \mathbb{R}^d$
such that 
$$x=\sum_{i=0}^n t_i v_i,$$ 
where $t_i\in\mathbb{R}$ when $0\leq i\leq n$, and $\sum_{i=0}^n t_i=1$.
Besides, $v_0,\dots , v_n$ are called the vertices of $\sigma$, the number $n$ is called the dimension of $\sigma$, and any simplex spanned by a subset of $\{v_0,\dots, v_n\}$ is called a face of $\sigma$.
\end{definition}

When a set of $n$-simplices is glued, a simplicial complex is formed.

\begin{definition}[simplicial complex]
A simplicial complex $K$ in $\mathbb{R}^d$ is a collection of simplices in  $\mathbb{R}^d$ such that:
\begin{enumerate}
    \item Every face of a simplex of $K$ is in $K$;
    \item the intersection of any two simplexes of $K$ is a face of each of them.
\end{enumerate}
Any $L\subset K$ is called a subcomplex of $K$ if $L$ is a simplicial complex.
\end{definition}

Next, the definition of $n$-chains and their boundaries is recalled. It is a key tool to formalize the idea of {\it holes} in a multidimensional space.

\begin{definition}[chain complexes]
Let $K$ be a simplicial complex and $n$ a dimension. An $n$-chain is a formal sum 
$c=\sum_{i=1}^{m} a_i\sigma_i$, 
where $\sigma_i$
are $n$-simplices of $K$ and $a_i\in \mathbb{Z}_2$ are
coefficients. 
The sum between $n$-chains is defined componentwise, i.e., let $c'=\sum_{i=1}^{m} b_i\sigma_i$ be another $n$-chain, then $c+c'=\sum_{i=1}^{m} (a_i+b_i)\sigma_i$. The $n$-chains together with the addition form an abelian group denoted by $C_n$. 
To relate these groups with different dimension, the boundary of an $n$-simplex $\sigma = \{v_1,\dots,v_n\}$ is defined
as the sum of its $(n-1)$-dimensional faces, 
that is,
$$\partial_n \sigma = \sum_{j=0}^n \{v_0,\dots, \hat{v}_j,\dots, v_n \},$$ 
where the hat on $v_j$ indicates that $v_j$ is omitted. 
The boundary of an $n$-chain is the sum of the boundaries of its simplices. Hence, the boundary operation $\partial_n$ is a homomorphism that maps an $n$-chain to an $(n-1)$-chain. 
Then, a chain complex is the sequence of chain groups connected by boundary homomorphisms,
$$\dots \xrightarrow{\partial_{n+2}}C_{n+1}\xrightarrow{\partial_{n+1}}C_n\xrightarrow{\partial_n}C_{n-1}\xrightarrow{\partial_{n-1}}\dots $$
\end{definition}

Next, the chains with empty boundary are considered. From an algebraic point of view, they have a group structure.

\begin{definition}[$n$-cycles and $n$-boundaries]
The group of $n$-cycles is the subgroup of the group of $n$-chains denoted by $Z_n$ composed of those $n$-chains $c$ with empty boundary, that is, $\partial_n c = 0$. 
The group of $n$-boundaries is the subgroup of the group of $n$-chains denoted by $B_n$ composed of those chains that are in the image of the $(n+1)$-st boundary homomorphism, that is,  $B_n=\im\,\partial_{n+1}$.
\end{definition}
Let us observe that since $\partial_{n+1}\partial_{n}=0$ then $B_n$ is a subset of $Z_n$. Therefore, we can already recall the definition of homology groups that determine the holes in the underlying space of a simplicial complex.

\begin{definition}[homology groups]
The $n$-th homology group is the quotient of the $n$-boundaries over the $n$-cycles, that is, 
$H_n=Z_n/B_n$.
The elements of $H_n$ are called $n$-homology classes.
\end{definition}

Next, we recall how to build a nested sequence of simplicial complexes in order to track the evolution of the
homology groups throughout the sequence. 

\begin{definition}[sublevel set filtration]
Given a simplicial complex $K$ and a continuous increasing
function $f: K \rightarrow \mathbb{R}$ called {\it filter function,} the {\it sublevel set filtration} ${\cal K}$
is a nested sequence  of subcomplexes of $K$ defined as:
$${\cal K}=\{K(a)= f^{-1}(-\infty, a]: a\in \mathbb{R}\}.$$
\end{definition}

Let us observe that $K(a)\subseteq K(b)$ when $a \le b$ since $f$ is increasing.
The sublevel set filtration that we will use in this paper is the so-called {\it Vietoris-Rips filtration,} which is a filtration  usually applied to point clouds. 
The Vietoris-Rips filter function enlarges $n$-balls from each point in the point cloud. 
Then, when two of these $n$-balls intersect, a $1$-simplex is built.
The process is extrapolated to higher dimensions, that is, if three balls intersect, a $2$-simplex is built, and so on.

As previously mentioned, in general, for every $a\leq b$, an inclusion map from $K(a)$ to $K(b)$ is considered.
Therefore, we have an induced homomorphism $f_n^{a,b}$ from  
$H_n(K(a))$ to $H_n(K(b))$.

\begin{definition}[persistent homology]
The sequence of $n$-th homology groups connected by homomorphisms obtained from a filtration ${\cal K}$ is called the $n$-th persistent homology of ${\cal K}$.
\end{definition}

Now, using the homology homomorphisms induced by the inclusion maps, we say  that an $n$-homology class $\alpha$ was born at $H_n(K(a))$ if it is not the image of any $n$-homology class  $\alpha'\in H_n(K(a'))$ with $a'<a$ and $f_n^{a'a}(\alpha')=\alpha$. 
It dies entering $H_n(K(b))$, with $a\leq b$, if it is the image of $\alpha$ and it is the image of another class born earlier than $\alpha$.
If $b-a$ is ``close'' to 0, then $\alpha$ is considered to be noise.  

\begin{definition}[persistence diagrams]
Let $\mu_n^{a,b}$ denote
the number of $n$-homology classes born at $H_n(K(a))$ and dying entering $H_n(K(b))$.
Then, the $n$-th persistence diagram of a filtration ${\cal K}$, denoted by $\Dgm_n({\cal K})$, is the multiset of points 
$(a,b)$ with multiplicity $\mu_n^{a,b}$ (together with the points of the diagonal with infinity multiplicity by convention).
\end{definition}

\begin{figure}
         \centering
         \includegraphics[width=0.32\textwidth]{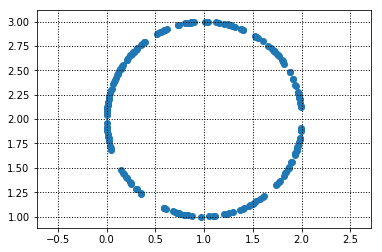}
         \includegraphics[width=0.32\textwidth]{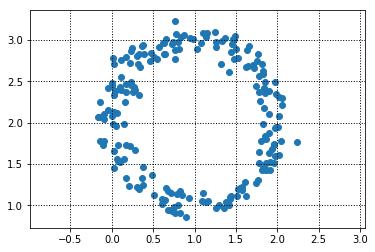}
         \includegraphics[width=0.32\textwidth]{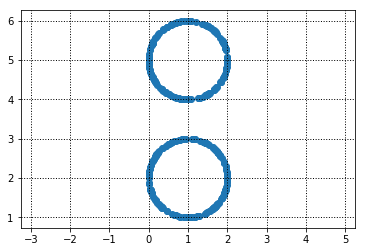}
         \caption{From left to right: A 2-dimensional point cloud sampling a circumference, a 2-dimensional point-cloud sampling a noisy circumference, and a 2-dimensional point-cloud sampling two circumferences.}
         \label{fig:circum_datasets}
\end{figure}

\begin{figure}
         \centering
         \includegraphics[width=0.32\textwidth]{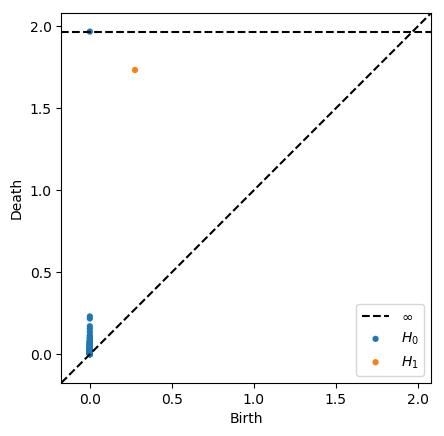}
         \includegraphics[width=0.32\textwidth]{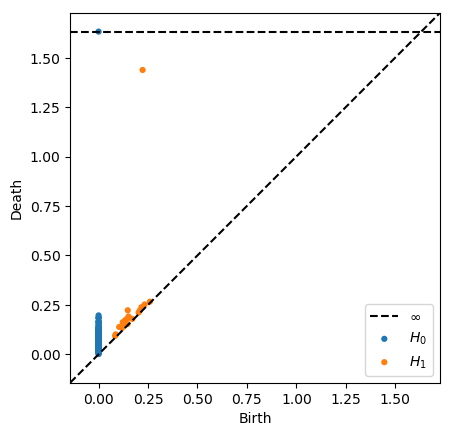}
         \includegraphics[width=0.32\textwidth]{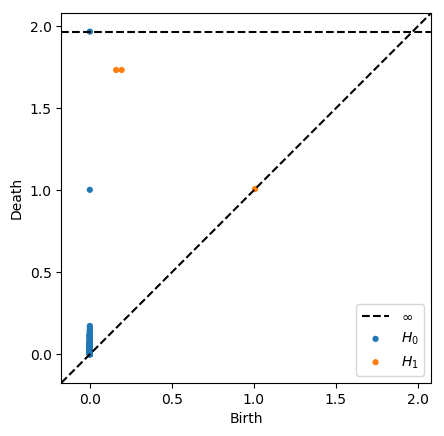}
         \caption{Three persistence diagrams of the Vietoris-Rips filtration obtained from a dataset (see Figure~\ref{fig:circum_datasets}) of a random selection of points belonging to a circumference and from two circumferences, respectively, with the 0- and 1-homology classes. 
         The blue points represent the {\it (birth,death)} of the 0-homology classes, and the orange points are the {\it (birth,death)} of the 1-homology classes. 
         Observe that in the third persistence diagram there are two orange points that are far from the diagonal corresponding to the
         holes of the two circumferences.}
         \label{fig:pers_diagrams}
     \end{figure}

Let us describe now a toy example as an illustration of the concept of persistent diagrams. Let us consider the 
three different datasets shown in Figure \ref{fig:circum_datasets}. The first one samples a circumference, the second one samples a noisy version of a circumference, and the last one samples two circumferences. 
Vietoris-Rips filtration using the Euclidean metric was computed to obtain the persistence diagrams shown in Figure \ref{fig:pers_diagrams}. 
The 2-dimensional blue and orange points of the persistence diagrams correspond, respectively,  to the
0- and 1-homology classes with birth and death time values being the coordinates. 
Looking at the persistence diagram showed  in Figure \ref{fig:pers_diagrams} on the left, we can observe  that 
just one significant
1-homology class is presented that corresponds to the {\it hole} of the circumference. 
However, in the persistence diagram showed in Figure \ref{fig:pers_diagrams} on the center, the points that appear close to the diagonal can be considered noise. 
Finally, the two orange point of the persistence diagram showed in Figure \ref{fig:pers_diagrams} on the left,  correspond to the two {\it holes}, one for each circumference sampled by the dataset displayed in Figure~\ref{fig:circum_datasets} on the right. 

To summarize the information of a persistence diagram, we will make use of the  persistent entropy concept \cite{CHINTAKUNTA2015391,e17106872} which has been proven to be stable under small perturbations in the input data (see  \cite{DBLP:journals/jiis/AtienzaGR19}).

\begin{definition}[persistent entropy]
Given a filtration $\mathcal{K}$ and the corresponding persistence diagram $\Dgm_n(\mathcal{K}) =\{(x_i,y_i)\ \mid \ 1\le i \le n\}$ (seen as a finite set of points), the $n$-th persistent entropy of $\mathcal{K}$  is defined as
$$\Ent_n(\mathcal{K})=-\sum_{i=1}^{n} p_{i} \log \left(p_{i}\right),$$
where $p_i=\frac{\ell_i}{L}$, $\ell_i=y_i-x_i$, and $L=\sum_{i=1}^n \ell_i$.
\end{definition}

Let us remark that those homology components that do not die (blue points in the horizontal dot line in Figure~\ref{fig:pers_diagrams}) will not be considered for the persistent entropy computation. For example, the persistent entropy values of the $0$-th persistence diagrams plotted in Figure~\ref{fig:pers_diagrams} are, from left to right, $4.58$, $4.49$, and $4$.

Finally, two persistence diagrams can be compared using a distance, the bottleneck distance being  considered the most common and the one that will be used in the next sections.

\begin{definition}[bottleneck distance]
The bottleneck distance between two persistence diagrams $A$ and $B$ is:
$$d_b(A,B)= \inf_{\phi :A\rightarrow B} \;\sup_{\alpha\in A} || \alpha-\phi(\alpha) ||_{\infty} $$
where $\phi$ is any possible bijection between $A$ and $B$.
\end{definition}

\begin{figure}
    \centering
    \includegraphics[width = 0.3 \linewidth]{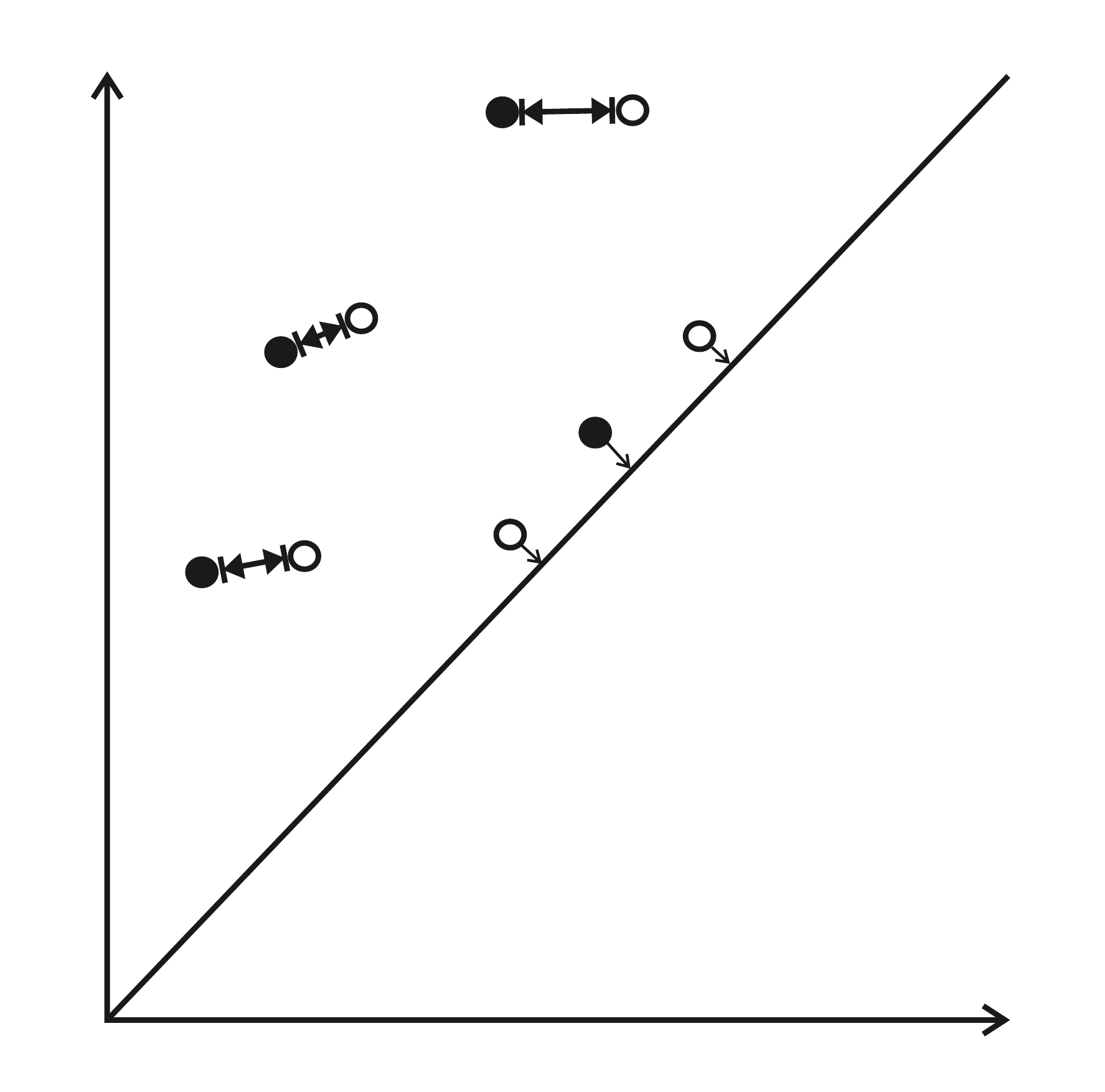}
    \caption{
     The set of arrows represents the
   optimum bijection between the black and white points that belong, respectively, to two different persistence diagrams, which are shown overlaid here.}
    \label{fig:bottleneck_example}
\end{figure}

A graphical description of the bottleneck distance is shown in Figure \ref{fig:bottleneck_example}.

\section{Computing TDA-based Summaries  of Sets of Texts
and Quantitative Comparison between Them}\label{sec:methodology}

Next, we describe the methodology based on TDA techniques designed to compute a TDA-based summary feature to any given set of texts and to establish a distance between
different sets of texts. 
In the next section, we will see that the TDA-based summary  characterize the literature works of an author and the proposed distance can establish which authors' styles are ``closer". This way, our results will support the qualitative philological studies previously made.

Broadly speaking, given a corpus composed of texts belonging to different categories (e.g., authors, styles) a stemming process (which we call \texttt{stem}) is applied to each
text where the non-informative words (also called stop-words) are deleted. 
Then, the  \texttt{skipgram} word embedding $E$ 
(described in Section \ref{sec:background_word2vec}) is applied to the vocabulary of the corpus, obtaining
a high-dimensional representation of the words as a point cloud. 
The point cloud is divided in (overlapping) subsets with points (words) belonging to the same category (e.g., authors, styles). 
For each of these point clouds, the Vietoris-Rips filtration is computed to obtain the corresponding persistence diagrams and the persistent entropy, which constitutes the TDA-based summary of the category. 
The pseudocode of the  methodology 
explained above to compute a TDA-based summary of a set of texts and a quantitative comparison bewteen them  is shown in Algorithm~\ref{algo:methodology1}.

\begin{algorithm}
\caption{TDA-based summaries of  sets of texts and quantiative comparison between them.}
\label{algo:methodology1}
 \hspace*{\algorithmicindent} {\bf Input:} A set ${\cal T}=\{T_1,\dots,T_m\}$ where each $T_i$ is a set of texts.\\
\hspace*{\algorithmicindent} {\bf Output:} TDA-based summaries
$\{\Ent_0({\cal K}_1),\dots,\Ent_0({\cal K}_m)\}$\\
\hspace*{\algorithmicindent} \hspace{1.5cm} and bottleneck distance $\{d_{i,j}\}_{1\leq i<j\leq m}$.\\
\begin{algorithmic}[1]
\For {$i\in \{1,\dots,m\}$} 
\State $C_i=\{\text{word} \mid \ \text{word} \in \text{any text of } T_i, \text{word}\not \in \text{stop-words}\} \text{ (ordered set)}$
\EndFor
\State $C=\bigcup_{i=1}^m C_i$
\State $E=E(C)\in \mathbb{R}^d$ \text{(word embedding)}\;
\For {$j\in \{1,\dots,m\}$}
\State $ W_j=\{w\in E \ \mid \ w=E(\text{word}) , \ \text{word}\in C_j
		\}
		$\;
\State  ${\cal K}_j=$ Vietoris-Rips filtration of $W_j$
\State	$\Dgm_0({\cal K}_j)=$ 0-th persistence diagram of ${\cal K}_j$\;
\State $\Ent_0({\cal K}_j)=$ 0-th persistent entropy of $\Dgm_0({\cal K}_j)$.
\For{$i\in \{1,\dots,j-1\}$}
\State $d_{i,j}=d_b(\Dgm_0({\cal K}_i),\Dgm_0({\cal K}_j))$
			\;
\EndFor
\EndFor
\end{algorithmic}
\end{algorithm}

\section{Experiment}\label{sec:experiment}
In this section, we illustrate the methodology presented above
and describe thoroughly the experimentation process 
accomplished
on the literature works of three well-known Spanish Golden Age poets\footnote{The code is available at the link \url{https://github.com/Cimagroup/TDA-based-text-metrics}.}.
In order to determine if 
there exist  significant differences between the 
TDA-based summary of the literature works of the three poets,
we apply a non-parametric statistical test to the resultant persistent entropy values.
In the following subsections, we proceed to describe each of the steps of the experiment in detail.


\subsection{The Context: Spanish Golden Age Literature}

The Spanish Golden Age literature is a complex framework still alive in the sense that it remains an appealing subject for the literary experts. In this section, we will provide a justification borrowed from the literary experts that supports our experimental results, and recall the preliminary literary notions needed to understand them.

We are interested in 
studies related to what we 
consider the inner "stylistic configurations" of the sentences in order to capture them with the \texttt{word2vec} embedding. Following the study developed by D\'amaso Alonso \citep{alonso1966poesia}, poets draw on different stylistic configurations for their verses. The first one we would like to comment can be exemplified by the following two verses of a sonnet by Cervantes \cite{BVMC:239403}:
\begin{verse}
{\it Afuera el fuego, el lazo, el hielo y la flecha \\
de amor que abrasa, aprieta, enfr\'ia y hiere...}
\end{verse}
We can see that the main concepts
of the first verse correspond member by member to the ones of the second verse, summarizing the following four sentences in the two verses: {\it Afuera el fuego de amor que abrasa; afuera el lazo de amor que aprieta, afuera el hielo de amor que enfr\'ia, afuera la flecha de amor que hiere}. It can be described as the following formula:
\begin{equation*}
    \begin{split}
      \alpha(A_1 \ \dots \ A_n)\\
      \beta(B_1 \ \dots \ B_n)
    \end{split}
\end{equation*}
that summarizes the 
sentences $\alpha(A_i)\beta(B_i)$ for $i=1\dots n$. 
In this example,
$\alpha$ is {\it afuera} and $\beta$ is {\it de amor}. 
Other kind of resource is the reiterative correlation plurality described in depth in \citep{damversos}. 
These techniques illustrate the big panoply of methods
that concern the configurations of the verses, but many others could be cited.

Our aim with the \texttt{word2vec} algorithm is to encapsulate this kind of configurations. In spite of its intrinsic difficulty, our work explores the possibility of finding
similarities between words and their use taking into consideration 
their context.
It seems natural, in a first approach, to 
study if 
\texttt{word2vec} with its \texttt{skipgram} variation can imitate or be used  
as a complement to the qualitative
methods in order to distinguish different literature styles.
 Besides, looking at the mathematical formulation to study the architecture of the sonnets  introduced by D\'amaso Alonso
 and his comment\footnote{Free translation. Original comment in Spanish.} 
 {\it "it would be a labour of a truly team of workers"} to apply such deep studies, in this paper, we take the chance to do that heavy work that D\'amaso Alonso mentioned, with recent mathematical tools in a efficient and effective way. 

Luis de G\'ongora and Lope de Vega are, both of them, summits of the 
Spanish Golden Age. Traditionally, it is said that Luis de G\'ongora started the literature style Culteranismo  and that Lope de Vega is related to an opposite trend called Conceptismo  which had its major representative in Francisco de Quevedo \citep{chamorro1987origenes, rutherford2016spanish}.
See also \citep{molfulleda2018oposicion} where it is claimed that both literary styles are related but with elements that distinguish them. However, there exists discrepancies between the literary experts.
For example, in \citep{alonso1966poesia}, D\'amaso Alonso did a thorough study of Lope de Vega, and he even developed a study of the comparison of this author with G\'ongora. He stated that there existed a discontinuous influence by the G\'ongora's work on the Lope de Vega's work. So, it might not be possible (and it is natural not to be so) to establish rigid difference between such literary styles.
In fact, poets present an evolution through their entire productive life, and the different literary styles can be inspired or fed by others. 
We also recommend \citep{Rozas} as an study of the context of these three poets. 

Hence, it is important to  highlight that the conclusions of our proposed technique can only be applied to the chosen sets of  texts and they can not be generalized to all the production of an author.


\subsection{The Corpus and the Preprocessing Step}\label{sec:corpstem}

The corpus we used is a huge dataset\footnote{The dataset can be found in \url{https://github.com/bncolorado/CorpusSonetosSigloDeOro}} composed of the sonnets from the Spanish Golden Age poets \citep{corpus}. 
It provides some metrical annotations according to stressed syllables, type of rhyme, etc. In our case we used the sonnets of the three poets we are interested in: Lope de Vega, Quevedo, and G\'ongora. 

Since, in the database, there are only 115 sonnets of G\'ongora,
we kept 115 sonnets of each poet (345 
sonnets in total) in order to avoid an unbalanced dataset. 
We chose just the first 115 sonnets of each poet 
in
the cited dataset,
without taking into consideration
any possible classification
that the literary experts could consider.

Then, each sonnet was pruned as a result of a stemming process. There exists some words that have no value in terms of meaning or that do not provide structure to the sentence such as prepositions: {\it de, el, la, ...} As they can be considered noise to the aim we follow, we erased them from the sonnets. Besides, some words are shortened to their
root in order to prevent
the \texttt{word2vec} algorithm from thinking
that different verb tenses or words with different genre are different words.
The procedure we applied to delete this non-informative words (also called stop-words) is implemented in the \texttt{nltk} library\footnote{\url{https://www.nltk.org/}}.


\subsection{Application of the \texttt{word2vec} Algorithm}\label{sec:word2vecapp}

This step consists 
in 
the application of the \texttt{skipgram} variation of the \texttt{word2vec} algorithm. 
Specifically, 
we used
the implementation 
provided by the \texttt{nltk} Python library.  
We then obtained a high-dimensional embedding of the words of the
345 
sonnets. 
Specifically,
the sonnets were embedded in a 150-dimension\-al space after a 250-iteration training using a window of 10 words. 
We used a window of 10 words because 
we wanted to catch patterns using the verses in their full extension, and 10 words is an 
upper bound to the number of words of a verse in a sonnet.

\subsection{Persistent Entropy
and 
Bottleneck Distance Computation}\label{sec:filtBot}

Having the high-dimensional representation of the words that compose the different sonnets of the dataset, we compute
the Vietoris-Rips filtration.
The metric used to compute
the Vietoris-Rips filtration is the cosine distance because it measures similarity between words by the angle of their vectors, and it is the common distance applied in
the \texttt{word2vec} algorithm (see \citep{Mikolov:2013:DRW:2999792.2999959}). 

As a result, we obtained 3 different 0-th persistence diagrams,
one for each poet. Then, the persistent entropy for each persistence diagram and the bottleneck distance between
any two persistence diagrams were computed.

\subsection{Results}
The methodology shown in Algorithm~\ref{algo:methodology1}
with the specific procedures and parameters described in Subsection \ref{sec:corpstem}, Subsection \ref{sec:word2vecapp}, and Subsection~\ref{sec:filtBot}, was 
applied and repeated 100 times. 

The persistent entropy values obtained after the 100 repetitions
were compared using non-parametric statistical tests. 
The results of the statistical tests are shown in Table~\ref{table:kruskal},
supporting that there exists a significant difference between the three set of sonnets and, hence, between the authors.

\begin{table}[]
\caption{
Tests applied to see
if there is a significant difference between 
the persistent entropy values obtained for the literary works of the 3 poets considered.}
\label{table:kruskal}
\centering
\begin{tabular}{lccc}
\multicolumn{4}{c}{Kruskal-Wallis' test}                                                             \\ \hline
\multicolumn{1}{l}{Contrast}          & \multicolumn{1}{c}{Significant difference} & \multicolumn{1}{c}{Difference} & \multicolumn{1}{c}{+/- Límits} \\ \hline
\multicolumn{1}{l}{G\'ongora - Lope de Vega}    & \multicolumn{1}{c}{Yes}    & \multicolumn{1}{c}{-200,0}     & \multicolumn{1}{c}{29,369}     \\ \hline
\multicolumn{1}{l}{G\'ongora - Quevedo} & \multicolumn{1}{c}{Yes}    & \multicolumn{1}{c}{-100,0}     & \multicolumn{1}{c}{29,369}     \\ \hline
\multicolumn{1}{l}{Lope de Vega - Quevedo}    & \multicolumn{1}{c}{Yes}    & \multicolumn{1}{c}{100,0}      & \multicolumn{1}{c}{29,369}     \\ \hline
\end{tabular}
\vspace{0.51cm}
\centering
\begin{tabular}{lccc}
\multicolumn{4}{c}{Friedman's test}                                                                                                     \\ \hline
\multicolumn{1}{l}{Contrast}          & \multicolumn{1}{c}{Significant difference} & \multicolumn{1}{c}{Difference} & \multicolumn{1}{c}{+/- Límits} \\ \hline
\multicolumn{1}{l}{G\'ongora - Lope de Vega}    & \multicolumn{1}{c}{Yes}    & \multicolumn{1}{c}{-2,0}       & \multicolumn{1}{c}{0,33856}    \\ \hline
\multicolumn{1}{l}{G\'ongora - Quevedo} & \multicolumn{1}{c}{Yes}    & \multicolumn{1}{c}{-1,0}       & \multicolumn{1}{c}{0,33856}    \\ \hline
\multicolumn{1}{l}{Lope de Vega - Quevedo}    & \multicolumn{1}{c}{Yes}    & \multicolumn{1}{c}{1,0}        & \multicolumn{1}{c}{0,33856}    \\ \hline
\end{tabular}
\end{table}

\begin{figure}
    \centering
    \includegraphics[width = 0.4 \textwidth]{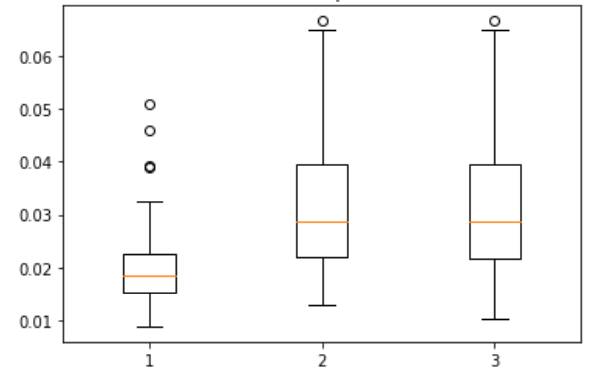}
    \caption{Box-plot showing the bottleneck distance results obtained from the sonnets of
    the three poets. (1) is the box-plot     of the bottleneck distance
    obtained from the comparison between the sonnets of
    Quevedo and Lope, (2) is the box-plot of the bottleneck distance obtained from the comparison between the sonnets of
    Quevedo and G\'ongora, and (3) is the box-plot of the bottleneck distance 
    obtained from the comparison between the sonnets of
    Lope de Vega and G\'ongora. 
    }
    \label{fig:boxplot}
\end{figure}

The bottleneck distances obtained after the 100 repetitions are shown in Figure \ref{fig:boxplot} using a box-plot representation. Let us recall that, in a box-plot, the higher horizontal line correspond to the maximum value and the lower horizontal line to the minimum value. The horizontal line in the middle of the box corresponds to the median, the top of the box is the third quartile, and the bottom of the box is the first~quartile. 
Finally, the circumferences correspond to outliers. 
We can see that the experimentation we applied can infer a significant difference between the bottleneck distances, being 
closer the persistence diagrams associated to the cloud points representing the literary works of Lope de Vega and Quevedo,  respectively.

\begin{table}
\caption{The repeated measures ANOVA applied to infer if there exists a significant difference between the bottleneck distances.  {\it Spher} means Sphericity assumed,
{\it G-G} means Greenhouse‑Geisser correction and
{\it H-F} means Huynh-Feldt correction.}

\begin{tabular}{ccccccc}
\hline
\multicolumn{2}{c}{\begin{tabular}{c}Source of\\ variation\end{tabular}}      & \begin{tabular}{c}Sum of\\squares\end{tabular}  & \begin{tabular}{c}Degree of\\freedom\end{tabular}      &\begin{tabular}{c} Mean\\square\end{tabular}  & \multicolumn{1}{c}{F-score}     & \multicolumn{1}{c}{$p$-value} \\ 
\hline
\multirow{3}{*}{Factor}   & Spher 
& 0.00834        & 2       & 0.00417     & \multicolumn{1}{c}{51.42} & \multicolumn{1}{c}{$<0.001$}  \\ \cline{2-7} 
                          & G-G
                          & 0.00834        & 1.126   & 0.00741     & \multicolumn{1}{c}{51.42} & \multicolumn{1}{c}{$<0.001$}  \\ \cline{2-7} 
                          &
                          H-F
& 0.00834        & 1.130   & 0.00738     & \multicolumn{1}{c}{51.42} & \multicolumn{1}{c}{$<0.001$}  \\ \hline
\multirow{3}{*}{Residual} &
Spher
& 0.0161         & 198     & 0.0000811   &                            &                                \\ \cline{2-5}
                          &
                         G-G 
                          & 0.0161         & 111.452 & 0.000144    &                            &                                \\ \cline{2-5}
                          & H-F
                         & 0.0161         & 111.850 & 0.000144    &                            &                                \\ \cline{1-5}
\end{tabular}
\label{tab:rmanova}
\end{table}
\begin{table}
\caption{Pairwise comparison between bottleneck distances. $A$, $B$, and $C$ correspond to the sample of the bottleneck distances between Lope de Vega and G\'ongora, Quevedo and G\'ongora, and Quevedo and Lope de Vega, respectively.}

\centering
\begin{tabular}{ccccccc}
\hline
\multicolumn{3}{c}{Factors}               & \begin{tabular}{c}Mean\\ difference \end{tabular}& \begin{tabular}{c}Standard\\ eError\end{tabular} & $p$-value                 & \begin{tabular}{c}95\% confidence\\ interval  \end{tabular}            \\ \hline
\multirow{2}{*}{$A$} & \multirow{2}{*}{with} & $B$ & -0.000386       & 0.000442   & 1.0000            & -0.00146 to 0.000690 \\ \cline{3-7} 
                   &                    & $C$ & 0.0110          & 0.00155    & \textless{}0.0001 & 0.00721 to 0.0148    \\ \hline
\multirow{2}{*}{$B$} & \multirow{2}{*}{with} & $A$ & 0.000386        & 0.000442   & 1.0000            & -0.000690 to 0.00146 \\ \cline{3-7} 
                   &                    & $C$ & 0.0114          & 0.00150    & \textless{}0.0001 & 0.00771 to 0.0150    \\ \hline
\multirow{2}{*}{$C$} & \multirow{2}{*}{with} & $A$ & -0.0110         & 0.00155    & \textless{}0.0001 & -0.0148 to -0.00721  \\ \cline{3-7} 
                   &                    & $B$ & -0.0114         & 0.00150    & \textless{}0.0001 & -0.0150 to -0.00771  \\ \hline
\end{tabular}
\label{tab:pair_comparison}
\end{table}

 Finally, in order to decide if the differences between the bottleneck distances computed are significant, a repeated measures ANOVA was applied\footnote{Statgraphics and MedCalc software (\url{https://www.medcalc.org/index.php}) were used to do the statistical validation of this section.}. 
 Sphericity is an assumption in repeated measures ANOVA designs. 
 When $\epsilon$ does not reach $1$, the $F$-score can be inflated and different corrections can be applied. 
 Then, in Table \ref{tab:rmanova}, 
 both corrections were applied as well with the sphericity assumption. 
 The Greenhouse-Geisser and Huynh-Feldt corrections, in case the sphericity assumption is violated, are  $\epsilon=0.563$ and $\epsilon=0.565$,  respectively. 
 Then, in Table~\ref{tab:rmanova} the different values obtained by the application of the repeated measures ANOVA are displayed. 
 A $p$-value lower than $0.001$ and a $F$-value of $51.42$ were reached. 
 So, we can say that there exists a significant difference. 
Therefore, we can infer that there exists a significant difference between the 3 groups of bottleneck distances as we expected by visualizing Figure \ref{fig:boxplot}. 

Finally, to specifically determine which of the groups is the different one, a pairwise comparison was computed in Table~\ref{tab:pair_comparison}.
As it is shown, the $p$-value is lower than $0.001$ when we compare with $C$. Therefore, the sample of the bottleneck distances between Quevedo and Lope de Vega literary works is significantly different from the other two. The $p$-value and the confidence intervals were Bonferroni corrected.

We conclude that our method shows that the distance (on the studied sonnets) of Quevedo and Lope de Vega literary works is significantly shorter than their distances to G\'ongora literary work.
Hence, 
we have found quantitative justification that support the philologists' theory that Quevedo and Lope de Vega belong to the same literary style {\it (Conceptismo)} and both of them are {\it stylistically far} from Luis de G\'ongora (whose style belongs to the {\it Culteranismo}).

\section{Conclusion}
\label{sec:conclusions}
Extracting knowledge from
complex data\-sets is a hard work which requires the help of techniques coming from other fields.
In this sense,
representing the data as points in a metric space opens a bridge between research fields which are seemingly far apart. 
The use of 
TDA
techniques is a new research area which provides tools for comparing properties of point clouds in high-dimensional spaces, and therefore, for comparing the datasets represented by such point clouds.

In this paper, we propose the use of such TDA techniques in order to compare different literary styles.
In this approach, bottleneck distance between the persistence diagrams of the Vietoris-Rips filtration obtained from the cloud points representing sets of texts from
different writers encodes
the differences between
their literary styles
and quantifies the proximity between
them.

This novel approach opens a door for the interaction of TDA and philological research. TDA techniques can be applied in order to give a topological description of a work, a writer or an age and go deeper into their belonging to a greater trend.


\section*{Acknowledgements}
The work was partly supported by the Agencia Estatal de Investigación/10.13039/501100011033 under grant PID2019-107339GB-100 and the Agencia Andaluza del Conocimiento under grant PY20-01145.




\end{document}